\documentclass[conference]{IEEEtran}
\usepackage[utf8]{inputenc}
\usepackage{booktabs}
\usepackage{graphicx}
\usepackage{amsmath}
\usepackage{caption}
\usepackage{float}
\usepackage{array}
\usepackage{hyperref}

\begin{document}

\title{Movie Revenue Prediction\\ Using Machine Learning Models}

\author{
    \IEEEauthorblockN{Vikranth Udandarao} \vspace*{3.0pt}
    \IEEEauthorblockA{
        \textit{Computer Science \& Engineering Dept.} \\
        \textit{IIIT-Delhi, India} \\
        vikranth22570@iiitd.ac.in
    }
    \and
    \IEEEauthorblockN{Pratyush Gupta} \vspace*{3.0pt}
    \IEEEauthorblockA{
        \textit{Computer Science \& Engineering Dept.} \\
        \textit{IIIT-Delhi, India} \\
        pratyush22375@iiitd.ac.in
    }
}

\maketitle

\begin{abstract}
    In the contemporary film industry, accurately predicting a movie's earnings is paramount for maximizing profitability. This project aims to develop a machine learning model for predicting movie earnings based on input features like the movie name, the MPAA rating of the movie, the genre of the movie, the year of release of the movie, the IMDb Rating, the votes by the watchers, the director, the writer and the leading cast, the country of production of the movie, the budget of the movie, the production company and the runtime of the movie. Through a structured methodology involving data collection, preprocessing, analysis, model selection, evaluation, and improvement, a robust predictive model is constructed. Linear Regression, Decision Trees, Random Forest Regression, Bagging, XGBoosting and Gradient Boosting have been trained and tested. Model improvement strategies include hyperparameter tuning and cross-validation. The resulting model offers promising accuracy and generalization, facilitating informed decision-making in the film industry to maximize profits.
\end{abstract}

\IEEEpeerreviewmaketitle

\section{Introduction}
    \subsection{Motivation}
        Imagine you are a filmmaker or head of a movie production house and you have a big question: what makes a movie a blockbuster hit or a flop?
        
        You might think it depends on the star power of the actors, the vision of the director, the budget of the production, or the genre of the story.
        
        Or you might think it is simply the quality of the storytelling that captivates the audience and earns high ratings.
        But the answer is not straightforward or easy.
        
        There are many factors that influence the earnings of a movie, and the true combination of these factors has not been mastered yet.
        That’s why we have developed a machine learning model that reveals the most important factors for box office success by analyzing real data from a wide variety of movies produced around the world.
        With our model, filmmakers can make more informed decisions and optimize their movie production for maximum profit and popularity.

    \subsection{Rationale}
        We hypothesize that certain parameters hold more significance in predicting movie revenue than others. Specifically, we conjecture that the director's track record and the genre of the film carry substantial weight in this prediction model.
        
        Our observations suggest that despite lower IMDb ratings, action-oriented films often demonstrate strong performance at the box office. Conversely, genres such as comedy or emotional dramas, despite potentially higher IMDb ratings, may not achieve comparable revenue outcomes to their action counterparts. 
        
        These insights underscore the complex interplay between film attributes and audience preferences, prompting us to assign greater importance to factors like directorial history and genre classification within our predictive framework.

    \subsection{Overview}
        In this project, we follow a structured methodology to build and evaluate our predictive model. We first collect a large dataset of movies and their features from various sources and custom tailor the datasets to suit our needs.
        
        We then pre-process the data to handle missing values, outliers, and categorical variables. We perform data analysis to explore the data and understand its characteristics and relationships. We use descriptive statistics, inferential statistics, and data visualization techniques to gain insights into the data like using a graph to compare the accuracy of our model's performance of training and test data.
        
        We then select several machine learning algorithms that are suitable for regression tasks, such as decision trees and random forests. We train and test our models using cross-validation and compare their performance using metrics such as R-squared mean error and Mean Absolute Percentage Error. We also apply model improvement strategies such as hyper-parameter tuning, feature selection and regularization to enhance the accuracy and generalization of our models. The resulting model offers promising results and can be used to predict the revenue of any movie based on its features.

\section{Literature Review}
    In this section, we will see the definitions of Principal Component Analysis \textit{(PCA)}, Label Encoder, SelectKBest features, GridSearchCV, Train Test Split and provide some literature content on the models which we are going to use. We will also explain the evaluation metrics '\textit{R² score}' and '\textit{MAPE}'.
    
    \subsection*{\href{https://en.wikipedia.org/wiki/Principal_component_analysis}{Principal Component Analysis (PCA)}}
    PCA is a statistical procedure that uses an orthogonal transformation to convert a set of observations of possibly correlated variables into a set of values of linearly uncorrelated variables called principal components. This technique is used to emphasize variation and bring out strong patterns in a dataset.
    
    \subsection*{\href{https://scikit-learn.org/stable/modules/generated/sklearn.preprocessing.LabelEncoder.html}{Label Encoder}}
    Label Encoder is a utility method to convert categorical data into numerical data. It assigns each unique category in the data to an integer value, making the data more suitable for algorithmic processing.
    
    \subsection*{\href{https://scikit-learn.org/stable/modules/generated/sklearn.feature_selection.SelectKBest.html}{SelectKBest Features}}
    SelectKBest is a feature selection method in Scikit-Learn. It selects features according to the k highest scores of a specified scoring function. It's a way to select the '\textit{k}' best features in your dataset, where '\textit{k}' is a parameter you choose.
    
    \subsection*{\href{https://scikit-learn.org/stable/modules/generated/sklearn.model_selection.GridSearchCV.html}{GridSearchCV}}
    GridSearchCV is a library function that is a member of sklearn's model\_selection package. It helps to loop through predefined hyperparameters and fit your estimator (model) on your training set. In addition to that, you can specify the number of times for the cross-validation for each set of hyperparameters.
    
    \subsection*{\href{https://scikit-learn.org/stable/modules/generated/sklearn.model_selection.train_test_split.html}{Train Test Split}}
    The '\textit{train\_test\_split}' function from the `\textit{sklearn.model\_selection}` module is a utility that divides a dataset into randomized training and testing subsets. Each subset is distinct, meaning that no data point can be present in both subsets. This allows for the model to be trained on one subset of the data, and then validated on an entirely separate subset. In our case, we applied an 80/20 split on our dataset for training and testing our models.
    
    \subsection*{Models}
    \subsection{\href{https://en.wikipedia.org/wiki/Linear_regression}{Linear Regression}}
    Linear Regression is a statistical approach for modelling the relationship between a dependent variable and one or more independent variables.
    \subsection{\href{https://en.wikipedia.org/wiki/Decision_tree}{Decision Trees}}
    A Decision Tree is a decision support tool that uses a tree-like model of decisions and their possible consequences. It is one way to display an algorithm that only contains conditional control statements.
    \subsection{\href{https://en.wikipedia.org/wiki/Gradient_boosting}{Gradient Boosting}}
    Gradient Boosting is a machine learning technique for regression and classification problems, which produces a prediction model in the form of an ensemble of weak prediction models, typically decision trees.
    \subsection{\href{https://en.wikipedia.org/wiki/Bootstrap_aggregating}{Bagging}}
    Bootstrap Aggregating, often abbreviated as Bagging, is a meta-algorithm designed to improve the stability and accuracy of machine learning algorithms used in statistical classification and regression. It also reduces variance and helps to avoid overfitting.
    \subsection{\href{https://en.wikipedia.org/wiki/Random_forest}{Random Forests}}
    Random Forests is a learning method that operates by constructing a multitude of decision trees at training time and outputting the class that is the mode of the classes (classification) or mean prediction (regression) of the individual trees.
    
    \subsection{\href{https://en.wikipedia.org/wiki/XGBoost}{XGBoosting}}
    XGBoost is an optimized distributed gradient boosting library designed to be highly efficient, flexible and portable. It implements machine learning algorithms under the Gradient Boosting framework. The XGBoost library offers several advantages over traditional machine learning algorithms, including:
        \begin{itemize}
        \item Parallelization: XGBoost supports parallelization, which allows it to train models efficiently on large datasets by leveraging multiple CPU cores or GPUs.
        \item Regularization: XGBoost includes built-in regularization techniques, such as L1 and L2 regularization, to prevent overfitting and improve model generalization.
        \item Handling Missing Data: XGBoost can automatically handle missing data without the need for imputation, making it more robust and efficient.
        \end{itemize}
    
    \subsection*{Evaluation Metrics}
    
    \subsection*{\href{https://en.wikipedia.org/wiki/Coefficient_of_determination}{R² Score}}
        The R² score, also known as the coefficient of determination, is a statistical measure that shows the proportion of the variance for a dependent variable that's explained by an independent variable or variables in a regression model. It provides an indication of goodness of fit and therefore a measure of how well unseen samples are likely to be predicted by the model.
    
    \subsection*{\href{https://stephenallwright.com/interpret-mape/}{Mean Absolute Percentage Error Error (MAPE)}}
        The Mean Absolute Percentage Error (MAPE) is a statistical measure used to assess the accuracy of a forecasting method in predictive studies.It is the mean of all absolute percentage errors between the predicted and actual values.It provides an understanding of the prediction error in terms of the percentage of the actual values. A lower MAPE value indicates a better fit of the data.Also MAPE can be interpreted as the inverse of model accuracy, but more specifically as the average percentage difference between predictions and their intended targets in the dataset. For example, if your MAPE is 10\% then your predictions are on average 10\% away from the actual values they were aiming for.

\section{Dataset Selection}
     \subsection{Modification of Dataset Strategy}
        In our initial proposal, we constructed a bespoke dataset by integrating four distinct datasets: The \href{https://www.kaggle.com/datasets/danielgrijalvas/movies}{\textbf{Movies Industry Dataset}}, \href{https://www.kaggle.com/datasets/rakkesharv/imdb-5000-movies-multiple-genres-dataset}{\textbf{IMDb 5000 Movies Multiple Genres Dataset}}, \href{https://www.kaggle.com/datasets/carolzhangdc/imdb-5000-movie-dataset}{\textbf{IMDb 5000 Movies Dataset}}, and \href{https://www.kaggle.com/datasets/mitchellharrison/top-500-movies-budget}{\textbf{Top 500 Movies Budget}}. This integration was performed sequentially, resulting in a \href{https://github.com/Vikranth3140/Movie-Revenue-Prediction/blob/main/old%20datasets/final_dataset.csv}{\textbf{comprehensive dataset encompassing 7118 movies}} with input variables such as budget, director, genre, leading crew, and IMDb ratings, and the output variable being gross revenue. (Refer to the \href{https://github.com/Vikranth3140/Movie-Revenue-Prediction/blob/main/old%20datasets/README.md}{\textbf{README}} for a detailed explanation of the dataset construction process).
        
        However, during the progression of our project, we identified the need to diverge from our originally proposed datasets. We opted for a refined dataset derived solely from \href{https://www.kaggle.com/datasets/danielgrijalvas/movies}{\textbf{The Movies Industry Dataset}}, achieved by excluding entries with missing values. This pivotal decision was informed by a multitude of considerations, which are detailed in the following subsections.
    
    \subsection{Rationale Behind the Dataset Transition}
        The primary challenge encountered with our initially finalized dataset was its suboptimal performance with our predictive models, reflected in lower accuracy rates. This was attributed to our initial approach of selecting input variables without a thorough analysis of the available data, leading us to force-fit the datasets into a preconceived framework of revenue determinants, thus compromising the integrity of the final dataset. Additionally, the potential for confusion due to movies sharing similar titles necessitated a cautious approach to dataset merging, which presented its own set of complexities.
        
        Acknowledging these issues, we paused to re-evaluate our methodology. We delved into a deeper examination of the data surrounding movie revenue prediction and gained insights into structuring the dataset to train a model that is both comprehensive and robust in forecasting movie earnings.
    
    \subsection{Benefits of the Optimized Dataset}
        Our continued exploration for an ideal dataset that encapsulates the core of our research led us to "The Movies Dataset," which aligned perfectly with our criteria. We ensured the dataset's reliability by eliminating entries with null values before subjecting it to our predictive models. A significant advantage of our current dataset is its ability to address the limitations of our prior data. It includes additional input variables such as Year, Production Company, and Votes \textit{(alongside the IMDb score)}, which collectively enhance the accuracy of our machine learning model in predicting movie revenues. Moreover, sourcing the dataset from a single origin has eliminated the extraneous noise that previously arose from dataset amalgamation.

\section{Data Analysis}
   \begin{table}[h]
    \centering
    \captionsetup{justification=centering}
    \caption{Description of Movie Dataset Features}
    \label{tab:movie_dataset_features}
    \renewcommand{\arraystretch}{1.5}
    \begin{tabular}{>{\centering\arraybackslash}m{1cm} >{\centering\arraybackslash}m{6cm}}
        \toprule
        \textbf{Feature Name} & \textbf{Description} \\
        \midrule
        Name & The title of the movie \\
        Rating & The MPAA rating of the movie \\
        Genre & The genre of the movie \\
        Year & The year the movie was released \\
        Released & The release date of the movie \\
        Score & The IMDb rating of the movie \\
        Votes & The number of votes the movie received \\
        Director & The person who directed the movie \\
        Writer & The person who wrote the movie script \\
        Star & The main actor or actress in the movie \\
        Country & The country where the movie was produced \\
        Budget & The budget of the movie \\
        Company & The production company of the movie \\
        Runtime & The duration of the movie \\
        \bottomrule
    \end{tabular}
\end{table}

    In the data analysis phase, we will thoroughly examine the collected movie dataset to gain insights into its structure, distribution, and relationships between features and the target variable \textit{(movie earnings)}. We will utilize descriptive statistics, inferential statistics, and data visualization techniques to explore the data.
    
    \subsection{Descriptive Statistics}
        We computed summary statistics such as mean, median, standard deviation, minimum, maximum, and quartiles for numerical features like budget and ratings. This gave us a general understanding of the central tendency and spread of the data.
    
    \subsection{Inferential Statistics}
        We will conduct statistical tests to analyze relationships between different features and the target variable. For example, we may use correlation analysis to examine the strength and direction of linear relationships between numerical features and earnings.
    
    \section{Data Visualization}
    \begin{figure}[H]
        \centering
        \includegraphics[width=0.3\textwidth]{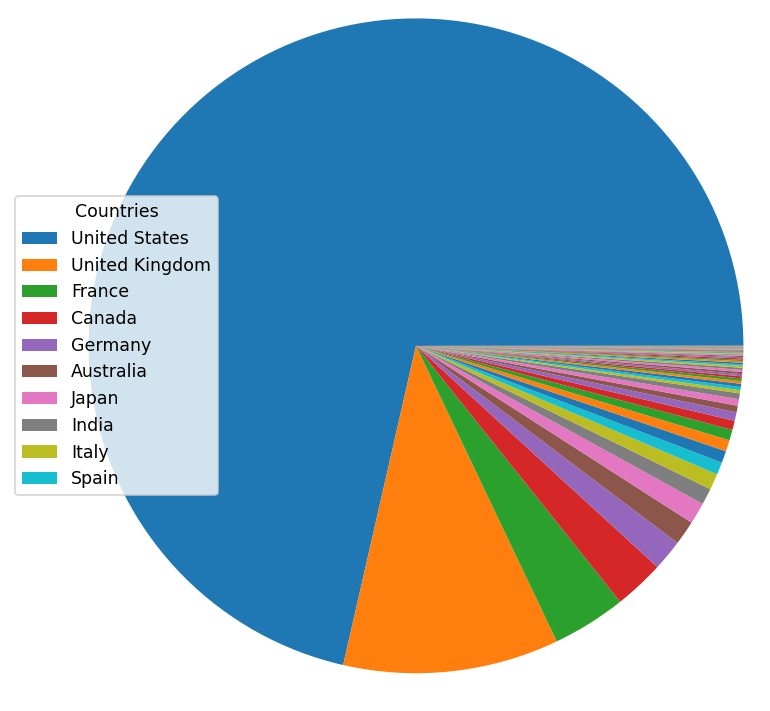}
        \caption{Distribution of Movies by Country}
        \label{fig:country-pie-chart}
    \end{figure}
    
    \begin{figure}[H]
        \centering
        \includegraphics[width=0.5\textwidth]{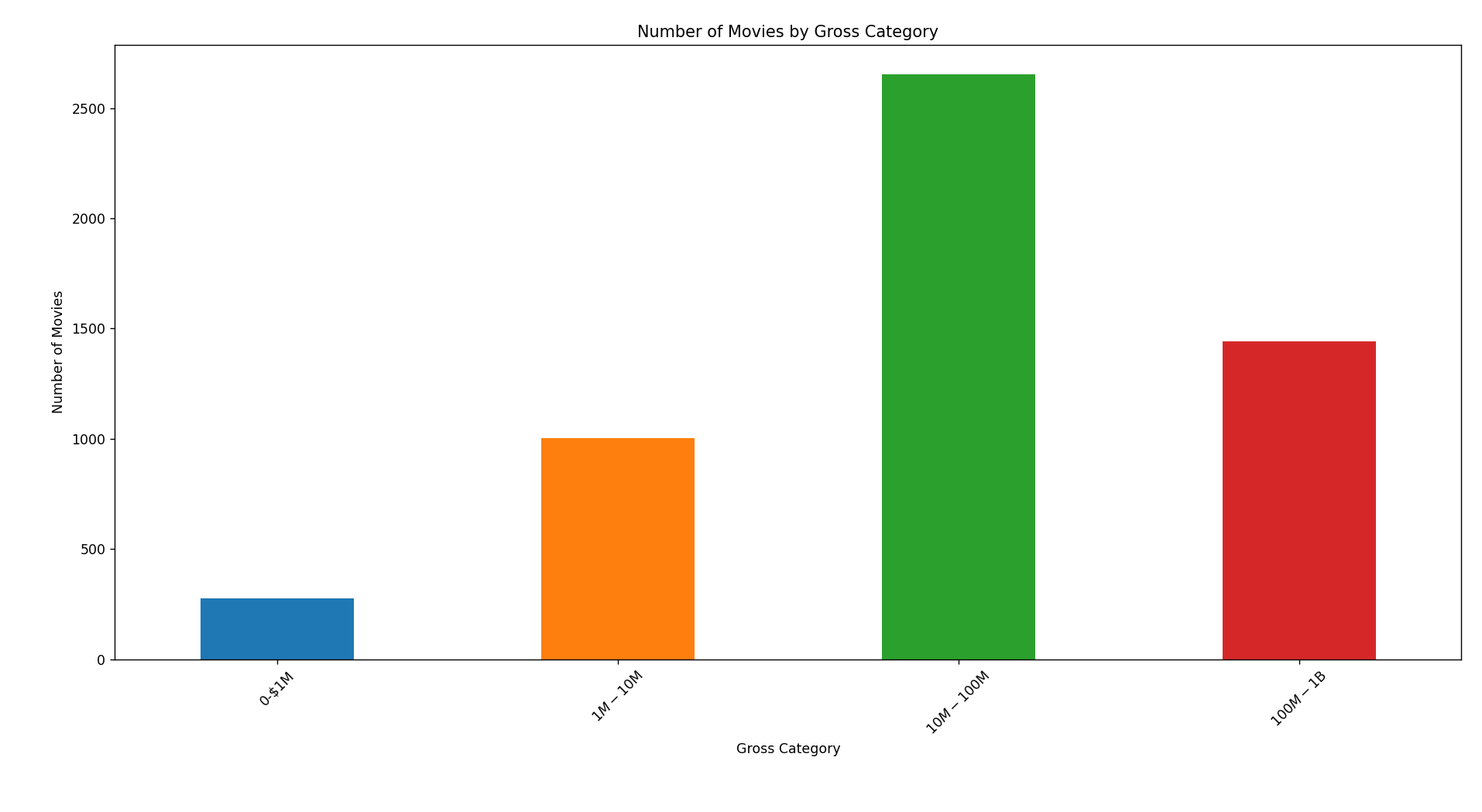}
        \caption{Histogram of Gross Categories}
        \label{fig:gross-histogram}
    \end{figure}

\section{Preprocessing}
    \subsection{Dataset Parameters}
        We have 14 parameters, including both non-numerical and numerical data types.
        
        \begin{enumerate}
            \item name
            \item rating
            \item genre
            \item year
            \item released
            \item score
            \item votes
            \item director
            \item writer
            \item star
            \item country
            \item budget
            \item company
            \item runtime
        \end{enumerate}
        
        We have 9 non-numerical data types
        \begin{enumerate}
            \item name
            \item rating
            \item genre
            \item released
            \item director
            \item writer
            \item star
            \item country
            \item company
        \end{enumerate}
        
        To prepare the data for regression models, we need to encode the non-numerical features into numerical labels. We will utilize the \href{https://scikit-learn.org/stable/modules/generated/sklearn.preprocessing.LabelEncoder.html}{LabelEncoder} from scikit-learn for this purpose.
        
        \subsection*{LabelEncoder}
            The LabelEncoder is a handy tool for normalizing and transforming non-numerical labels into numerical equivalents. It's important to note that the labels should be hashable and comparable for this process to work effectively. We will begin by fitting the label encoder and obtaining the encoded labels for further processing.
            
        \subsection*{Handling Null values}
            In the \href{https://www.kaggle.com/datasets/danielgrijalvas/movies}{\textbf{Movie Industry Dataset}}, there are 2,247 null values across the 11 parameters totalling 7669.
            Since budget and gross are our main parameter and output, we dropped those datasets and we were left with 5422 datasets.
            
            \begin{figure}
                \centering \includegraphics[width=0.4\linewidth]{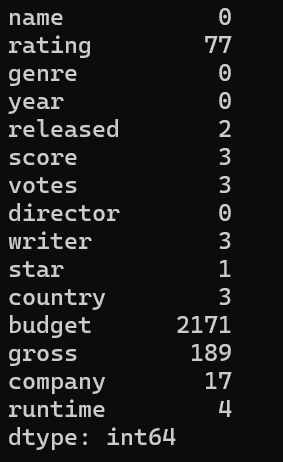}
                \caption{Null Values}
                \label{fig:null-values}
            \end{figure}
            
        \subsection*{PCA}
            We tried to use Principal Component Analysis but we were not able to achieve significant increase in the accuracy.
            We believe the reason for this is that PCA works effectively when there are more parameters and datasets.
            It will be more useful in image or sound regression/classification as the dimensions of the image/sound will come into play.
            Thus, we have decided to drop PCA in our preprocessing.
            
        \subsection*{Best Features}
            We wanted to know which features contribute more to the revenue of the movie.
            
            So we had used \href{https://scikit-learn.org/stable/modules/generated/sklearn.feature_selection.SelectKBest.html}{SelectKBest} and then categorized and predicted all the 8668 features which can be viewed in \href{https://github.com/Vikranth3140/Movie-Revenue-Prediction/blob/main/Helper%20files/Best%20Festures/feature_scores.txt}{\textbf{Features Scores}}.
            
            Since there were a lot of features for us to analyse, we then printed only those features with scores greater than 100 which can be viewed in \href{https://github.com/Vikranth3140/Movie-Revenue-Prediction/blob/main/Helper%20files/Best%20Festures/significant_features.txt}{\textbf{Best Features Scores}}.
            
            \begin{figure}[H]
                \centering
                \includegraphics[width=0.5\linewidth]{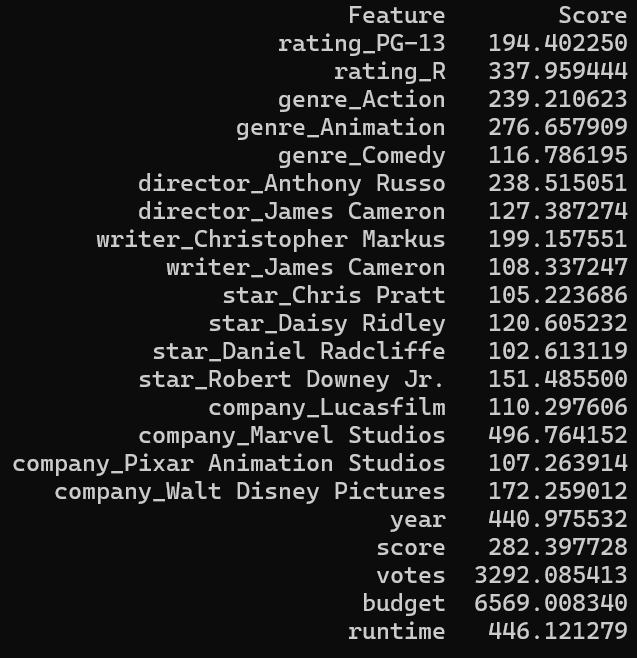}
                \caption{K Best Features}
                \label{fig:k-best}
            \end{figure}
            
            As per our intuition, the budget played was the most deciding feature, with a score of about 6569.
            However, to our surprise, votes played the next most important role in predicting the revenue.
            
            Additionally, runtime, year, and score also played a decent role in predicting the revenue.
            
            Also, there were a few individual companies and individuals in the field of direction, cast and production who had an impact on the way the movies performed at the box office.
            
            Also, there is a general trend that can be noticed where PG-13 and R-rated movies tend to do well.
            
            Action, Animation, and Comedy are well-performing genres which also tend to increase the revenue of the movie.

\section{Model Selection}
    For predicting movie earnings, we have explored several regression techniques and ensemble methods like Linear Regression with and without PCA Preprocessing, Decision Trees, Gradient Boosting, Bagging, Random Forests and XGBoosting.
    Ensemble methods like Random Forests and Gradient Boosting are capable of handling complex interactions and non-linearities in the data.
    
    \subsubsection{Linear Regression}
        Initially used to explore the dataset's linear relationships and determine if a straight-line model could adequately represent the data.
    
    \subsubsection{Gradient Boosting}
        Employed as a predictive technique to create an ensemble model consisting of weaker prediction models (typically decision trees). Each subsequent learner in the ensemble aims to correct errors made by its predecessors. GB is known for its robustness against overfitting, making it a valuable algorithm for predictive tasks.
    
    \subsubsection{Random Forest}
        Utilized as an ensemble learning method for classification. RF constructs multiple decision trees and makes predictions based on the majority vote of these trees. Despite efforts to reduce dataset dimensionality, the RF model exhibited overfitting on the training set and did not outperform other models significantly.
    
    \subsubsection{Decision Tree}
        Employed as a hierarchical decision support model. DTs use a tree-like structure to represent decisions and their consequences, including chance events, resource costs, and utility. They provide a clear visualization of the algorithm's decision-making process.
    
    \subsubsection{Bagging}
        Used to improve model performance by training multiple models on random subsets of the original data. The results from these models are combined through a voting mechanism to make predictions, resulting in more accurate and robust predictions.
    
    \subsubsection{XGBoost}
        Applied as a popular machine learning algorithm under ensemble learning. XGBoost is effective for supervised learning tasks like regression and classification. It iteratively builds a predictive model by combining predictions from multiple individual models, often decision trees, to enhance prediction accuracy and model performance.

\section{Model Improvement}
    In our pursuit of enhancing the model's performance, we will employ several strategies. Let's break down the key components:
    
    \subsection{Standardizing Data with \href{https://scikit-learn.org/stable/modules/generated/sklearn.preprocessing.StandardScaler.html}{Standard Scaler}}
    
        To ensure consistent scaling across features, we'll utilize the \textit{Standard Scaler}. This technique standardizes each feature by subtracting the mean and dividing by the standard deviation. By applying this transformation, we bring all features to a similar scale, which can significantly improve the model's performance.
    
    \subsection{Feature Engineering: \href{https://numpy.org/doc/stable/reference/generated/numpy.log1p.html}{Logarithmic Transformations}}
    
        We'll focus on two specific columns: '\textit{budget}' and '\textit{gross}' revenue. These columns often exhibit skewed distributions. To address this, we'll apply logarithmic transformations. Converting '\textit{budget}' and '\textit{gross}' to \textit{\(\log(\text{budget})\)} and \textit{\(\log(\text{gross})\)} using \href{https://numpy.org/doc/stable/reference/generated/numpy.log1p.html}{\textbf{numpy.log1p}}, respectively, will stabilize their variance and enhance the model's robustness. We will also report MSLE when the above is implemented to check if the model's bias is low.
    
    \subsection{Hyperparameter Tuning using \href{https://scikit-learn.org/stable/modules/generated/sklearn.model_selection.GridSearchCV.html}{GridSearchCV}}
    
    Now, let's delve into the details of hyperparameter tuning. \textit{GridSearchCV} is a powerful tool that performs an exhaustive search over specified parameter values for an estimator. In our case, we're using a \textit{Gradient Boosting Regressor} with a squared error loss function. Here's how it works:
    
    \begin{itemize}
        \item We define a dictionary called \texttt{param\_grid} containing various hyperparameters (e.g., number of estimators, maximum depth, and learning rate).
        \item \textit{GridSearchCV} systematically tests different combinations of these hyperparameters using \href{https://scikit-learn.org/stable/modules/cross_validation.html}{cross-validation} (in our case, 5-fold cross-validation).
        \item The best combination of parameters, as determined by the highest R-squared score, is extracted using the \texttt{best\_params} attribute.
        \item Armed with these optimal parameters, we initialize the final model.
        \item The model is then trained on the training data and evaluated on both the training and test datasets to validate its accuracy.
    \end{itemize}
    
    By following this process, we aim to create a robust and high-performing model.

    \subsection{Implementing \href{https://xgboost.readthedocs.io/en/latest/python/python_api.html}{XGBRegressor} using \href{https://xgboost.readthedocs.io/en/stable/}{xgboost} library}
    XGBoost (Extreme Gradient Boosting) is a powerful and efficient implementation of the gradient boosting algorithm, which is widely used for various machine learning tasks, including regression and classification. In this project, we employed the XGBRegressor from the xgboost library to train and test a regression model for predicting movie revenues based on various features such as release date, genre, director, and more.
    
        To train and evaluate the XGBoost regression model, we followed these steps:
        \begin{enumerate}
        \item Load the movie dataset and preprocess the data by encoding categorical features using LabelEncoder from scikit-learn.
        \item Split the dataset into training and testing sets using \texttt{train\_test\_split} from scikit-learn.
        \item Define a custom callback class TrackR2Score that inherits from xgb.callback.TrainingCallback. This class overrides the after\_iteration method to calculate and store the R-squared score on the training set after each iteration during the training process.
        \item Perform hyperparameter tuning using GridSearchCV from scikit-learn. We passed the TrackR2Score callback to the XGBRegressor instance used as the estimator in GridSearchCV, allowing us to track the R-squared score during the grid search process.
        \item Train the final XGBRegressor model using the best hyperparameters found by GridSearchCV.
        \item Evaluate the model's performance on both the training and testing sets using various metrics, such as R-squared score and Mean Absolute Percentage Error (MAPE).
        \item Visualize the actual vs. predicted values for both the training and testing sets.
        \item Plot the training R-squared score curve to observe the gradual improvement of the model during the training process.
        \end{enumerate}
    The following plot shows the training R² score curve, which illustrates the model's gradual improvement during the training process:
    \begin{figure}[h]
        \centering
        \includegraphics[width=1\linewidth]{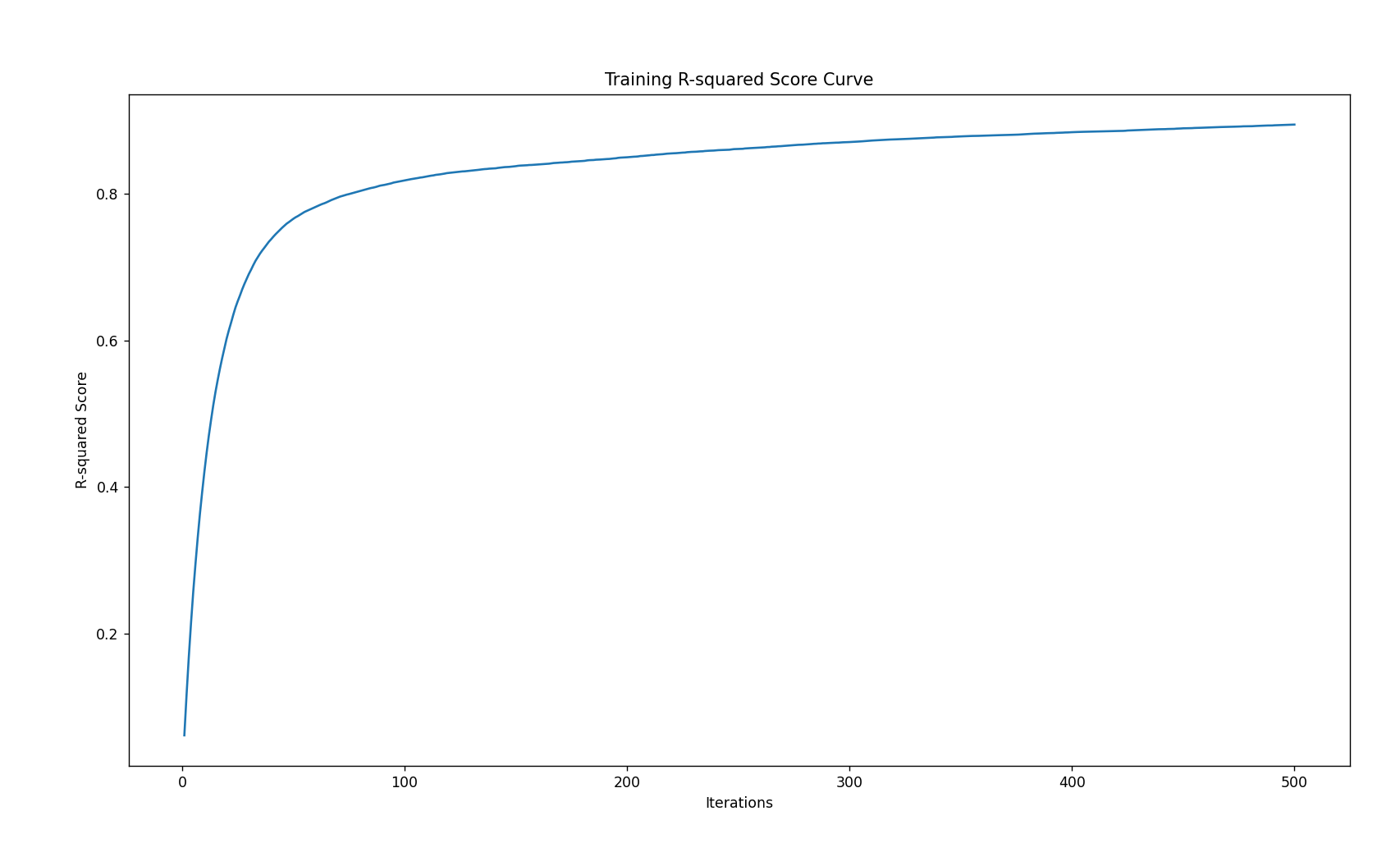}
        \caption{Training R² Score Curve}
        \label{fig:xg_boost_track}
    \end{figure}
    By leveraging the powerful XGBoost library and implementing a custom callback function, we were able to effectively train and evaluate a regression model for predicting movie revenues while monitoring the model's performance during the training process.

\section{Model Evaluation}
    We will evaluate the performance of our models using appropriate metrics such as Mean Absolue Percentage Error\textit{(MAPE)} and Coefficient of Determination \textit{(R²)}. Taken together, these two metrics provide a comprehensive view of your model’s performance. The \textit{R²} score tells you how well your model is capturing the patterns in the data, while the \textit{MAPE} gives you a sense of the average percentage error in your model’s predictions. Additionally, we will compare the performance of our model with baseline models to assess its effectiveness.
    
    \begin{table}[h]
        \centering
        \renewcommand{\arraystretch}{1.1}
        \begin{tabular}{
            | p{0.20\textwidth}
            | p{0.12\textwidth} 
            | p{0.12\textwidth} |
            | p{0.12\textwidth}
            }
            \hline
            \textbf{Model} & \textbf{R²} & \textbf{MAPE} \\
            \hline
            \multicolumn{1}{|l}{\textbf{Linear Regression}} \\
            \hline
            Training Set   & 0.6553 & 35.23\% \\
            \hline
            Testing Set & 0.6706 & 18.49\% \\
            \hline
            \multicolumn{1}{|l}{\textbf{Decision Tree}}\\
            \hline
            Training Set & 0.8664 & 13.00\% \\
            \hline
            Testing Set & 0.6947 & 4.60\%\\
            \hline
            \multicolumn{1}{|l}{\textbf{Bagging}} \\
            \hline
            Training Set & 0.8583 & 13.32\% \\
            \hline
            Testing Set & 0.7719 & 5.67\% \\
            \hline
            \multicolumn{1}{|l}{\textbf{Gradient Boost}}\\
            \hline
            Training Set & 0.9158 & 10.57\% \\
            \hline
            Testing Set & 0.8242 & 5.69\% \\
            \hline
            \multicolumn{1}{|l}{\textbf{eXtreme Gradient Boosting}}\\
            \hline
            Training Set & 0.9079 & 9.70\% \\
            \hline
            Testing Set & 0.8102 & 5.53\%\\
            \hline
            \multicolumn{1}{|l}{\textbf{Random Forest}}\\
            \hline
            Training Set & 0.8728 & 14.29\% \\
            \hline
            Testing Set & 0.7786 & 5.33\% \\
            \hline
        \end{tabular}
    \caption{Model Evaluation Result}
    \end{table}
    
    \begin{figure}[h]
        \centering
        \includegraphics[width=1\linewidth]{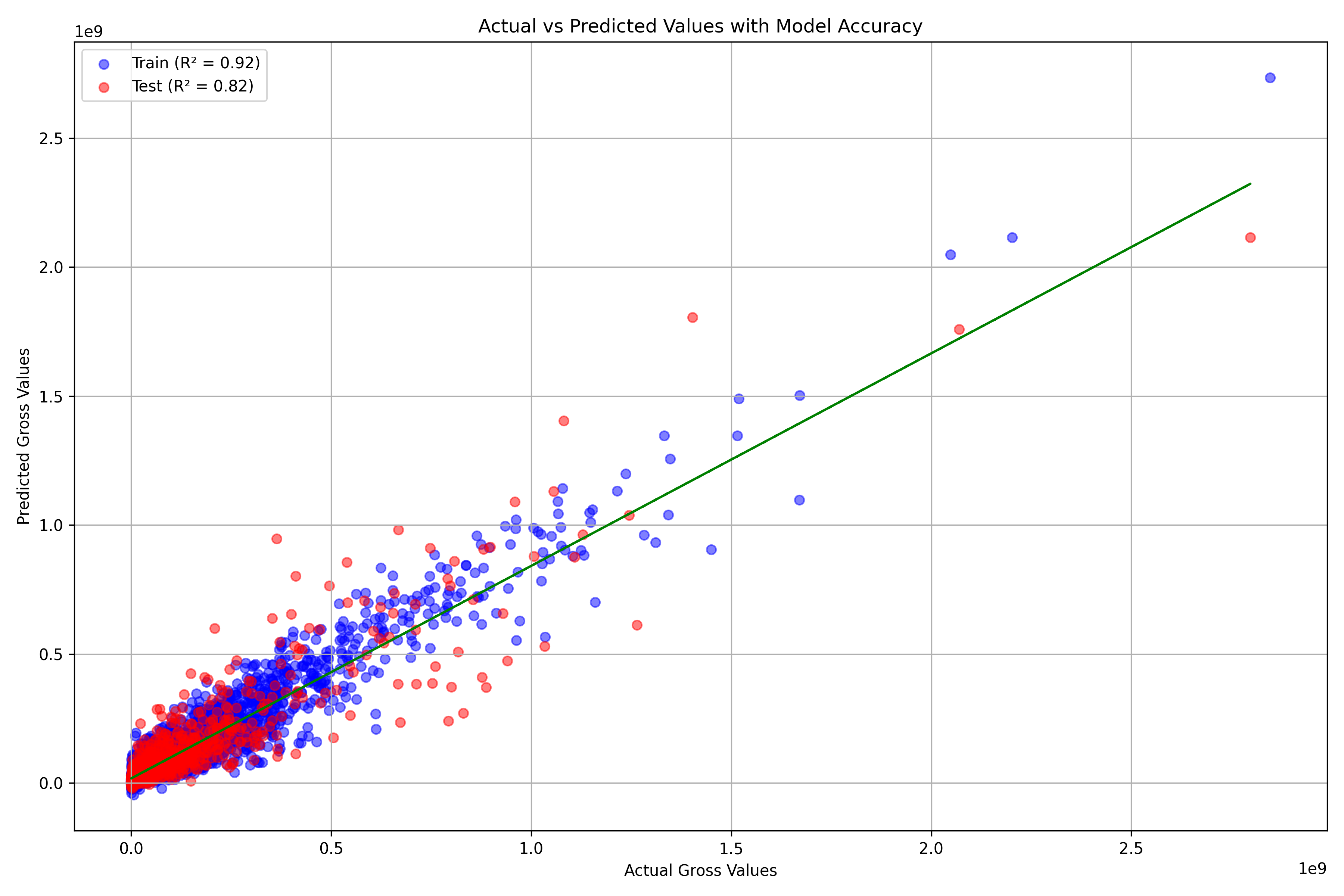}
        \caption{Gradient Boosting Test Results}
        \label{fig:gradient_boosting_test}
    \end{figure}

\section{Command Line Interface}
    We have made a command line interface for producers to access our model, where they will be prompted to input the 14 parameters on which the model will be trained.
    We have the option for the producer to choose the model which he wants to use.
    \begin{enumerate}
        \item Linear Regression
        \item Decision Tree
        \item Bagging
        \item Random Forest
        \item XGBoost
        \item Gradient Boosting
    \end{enumerate}

    The CLI can be accessed in \href{https://github.com/Vikranth3140/Movie-Revenue-Prediction/blob/main/main.py}{\textbf{main.py}}
    The producer has to run \textit{python main.py} and then enter the 14 parameters to predict the revenue for his/her upcoming movie as we have envisioned in our Problem Statement.

\section{Conclusion}
    As of now, we have found Gradient Boosting to be our best model, achieving:
    
    \begin{itemize}
        \item Final Training Accuracy: 91.58\%
        \item Final Testing Accuracy: 82.42\%
    \end{itemize}
    
    The application can be accessed in \href{https://github.com/Vikranth3140/Movie-Revenue-Prediction/blob/main/main.py}{\textbf{main.py}}
    
    The model can be viewed in \href{https://github.com/Vikranth3140/Movie-Revenue-Prediction/blob/main/models/gradient_boost.py}{\textbf{Models/gradient\_boost.py}}.
    
    The other models can be viewed \href{https://github.com/Vikranth3140/Movie-Revenue-Prediction/tree/main/models}{\textbf{here}}.
    
    The accuracies of all the models can be viewed \href{https://github.com/Vikranth3140/Movie-Revenue-Prediction/blob/main/models/accuracies.text}{\textbf{here}}.

\section*{Acknowledgment}
    The authors would like to extend their sincerest gratitude to \href{https://www.iiitd.ac.in/subramanyam}{Dr A V Subramanyam} \textit{(Computer Science \& Engineering Dept., \href{https://www.iiitd.ac.in/}{IIIT-Delhi})} for their invaluable guidance throughout the project.
    Their insightful feedback and expertise have been instrumental in shaping this project into its final form.


\begin{thebibliography}{1}
    \bibitem{ref1}
    Vr, Nithin \& Pranav, M \& Babu, PB \& Lijiya, A.. (2014). Predicting Movie Success Based on IMDB Data. International Journal of Business Intelligents. 003. 34-36. 10.20894/IJBI.105.003.002.004. 
    
    \bibitem{ref2}
    Pradeep, Kavya \& TintuRosmin, C \& Durom, Sherly \& Anisha, G. (2020). Decision Tree Algorithms for Accurate Prediction of Movie Rating. 853-858. 10.1109/ICCMC48092.2020.ICCMC-000158. 
    
    \bibitem{ref3}
    Garima Verma and Hemraj Verma, "Predicting Bollywood Movies Success Using Machine Learning Technique," IEEE, 2019.
    
    \bibitem{ref4}
    Rijul Dhir and Anand Raj, "Movie Success Predictions using Machine Learning and their Comparison," IEEE International Conference on Secure Cyber Computing and Communication, 2018.
    
    \bibitem{ref5}
    Ashutosh Kanitar, "Bollywood Movie Success Prediction using Machine Learning Algorithms," IEEE Third International Conference on Circuits, Control, Communication and Computing, 2018.
    
    \bibitem{ref6}
    Wales, Lorene. (2017). The Complete Guide to Film and Digital Production: The People and the Process. 10.4324/9781315294896. 
    
    \bibitem{ref7}
    Beyza Çizmeci and Sule Gündüz Öğüdücü, "Predicting IMDb Rating of Pre-release Movies with Factorization Machines Using Social Media," IEEE 3rd International Conference on Computer Science and Engineering, 2018.
    
    \bibitem{ref8}
    Steve Shim and Mohammad Pourhomayoun, "Predicting Movie Market Revenues Using Social Media Data," IEEE International Conference on Information Reuse and Integration, 2017.
    
    \bibitem{ref9}
    Nahid Quader and Md. Osman Gani, "A Machine Learning Approach to Predict Movie Box-office," Information Technology (ICCIT), December 2017.
    
    \bibitem{ref10}
    Beyza Çizmeci and Sule Gündüz Öğüdücü, "Predicting IMDb Ratings of Pre-release Movies with Factorization Machines Using Social Media," International Conference of Computer and Information Technology (ICCIT), 2017.
    
    \bibitem{ref11}
    Subramaniyaswamy V., and Vignesh Vaibhav M., "Predicting Movie Box Office Success using Multiple Regression and SVM," International Conference of Computer and Information Technology (ICCIT), 2017.
    
    \bibitem{ref12}
    M.H. Latif, H. Afzal, "Prediction of Movies Popularity Using Machine Learning Techniques," National University of Sciences and Technology, H-12, ISB, vol. 16, no. 8, pp. 127–131, 2016.
    
    \bibitem{ref13}
    D.A., Olubukola \& O.M., Stephen \& A.K., Funmilayo \& Omotunde, Ayokunle \& A., Oyebola \& Oduroye, Ayorinde \& Ajayi, Wumi \& Yaw, Mensah. (2021). Movie Success Prediction Using Data Mining. British Journal of Computer, Networking and Information Technology. 4. 22-30. 10.52589/BJCNIT-CQOCIREC. 

\end{thebibliography}
\end{document}